\documentclass[a4paper]{styles/svproc}

\usepackage{graphics}
\usepackage{color}
\usepackage{enumerate}
\usepackage{epsfig}
\usepackage{booktabs}
\usepackage{graphicx,url}
\usepackage{array}
\usepackage{latexsym}
\usepackage{amsfonts}
\usepackage{amsmath}
\usepackage{amssymb}
\usepackage{xstring}
\usepackage{multirow}
\usepackage{prettyref}
\usepackage{flexisym}
\usepackage{bigdelim}
\usepackage{breqn}
\usepackage{listings}
\usepackage{enumitem}
\usepackage{xspace}
\usepackage{makecell}
\usepackage[bottom]{footmisc}
\usepackage[table]{xcolor}
\usepackage{hyperref}
\hypersetup{
  colorlinks=true,
  citecolor=green!50!black
}
\usepackage{subcaption}
\makeatletter\@ifpackageloaded{subcaption}{}\makeatother

\newcommand{\eg}{\emph{e.g.,}\xspace}
\newcommand{\ie}{\emph{i.e.,}\xspace}

\newcommand{\ph}[1]{\textbf{#1.}}

\usepackage[bibstyle=ieee, citestyle=numeric-comp, backend=biber, mincitenames=1, maxcitenames=2, maxbibnames=20]{biblatex}
\usepackage{url}
\usepackage[acronym]{glossaries}

\begin{document}
\newacronym{tamp}{TAMP}{Task and Motion Planning}
\newacronym{dof}{DOF}{degree-of-freedom}
\newacronym{pddl}{PDDL}{Planning Domain Definition Language} \glsdisablehyper
\mainmatter

\title{Language-Grounded Hierarchical Planning and Execution with Multi-Robot 3D Scene Graphs}

\titlerunning{Language-Grounded TAMP via Multi-Robot 3DSGs}

\author{
Jared Strader\inst{1},
Aaron Ray\inst{1},
Jacob Arkin\inst{1},
Mason B. Peterson\inst{1},
Yun Chang\inst{1},
Nathan Hughes\inst{1},
Christopher Bradley\inst{1},
Yi Xuan Jia\inst{1},
Carlos Nieto-Granda\inst{2},
Rajat Talak\inst{1},
Chuchu Fan\inst{1},
Luca Carlone\inst{1},
Jonathan P. How\inst{1},
Nicholas Roy\inst{1}
}

\authorrunning{Strader et al.}

\institute{Massachusetts Institute of Technology, Cambridge, MA 02139, USA,\\
\and
U.S. Army Combat Capabilities Development Command, Army Research Laboratory, Adelphi, MD 20783, USA, \\
}

\maketitle

\renewcommand{\thefootnote}{}\footnotetext{\scriptsize
This work was sponsored by the Army Research Laboratory under Cooperative Agreement Number W911NF-17-2-0181, MIT Lincoln Laboratory’s Autonomy al Fresco program, and the ONR RAPID program.
}

\setcounter{footnote}{0}
\renewcommand{\thefootnote}{\arabic{footnote}}
\vspace{-3mm}
\begin{abstract}
In this paper, we introduce a multi-robot system that integrates mapping, localization, and task and motion planning (TAMP) enabled by 3D scene graphs to execute complex instructions expressed in natural language.
Our system builds a shared 3D scene graph incorporating an open-set object-based map, which is leveraged for multi-robot 3D scene graph fusion.
This representation supports real-time, view-invariant relocalization (via the object-based map) and planning (via the 3D scene graph), allowing a team of robots to reason about their surroundings and execute complex tasks.
Additionally, we introduce a planning approach that translates operator intent into Planning Domain Definition Language (PDDL) goals using a Large Language Model (LLM) by leveraging context from the shared 3D scene graph and robot capabilities.
We provide an experimental assessment of the performance of our system on real-world tasks in large-scale, outdoor environments.
A supplementary video is available at \url{https://youtu.be/8xbGGOLfLAY}.
\vspace{-4mm}
\end{abstract} \vspace{-4mm}
\section{Introduction}
\vspace{-2mm}
Imagine a team of robots tasked with carrying out complex operations specified by a human operator in a large-scale, outdoor environment.
Doing so may involve visually inspecting objects or regions of interest and interacting with the environment (\eg{} to clear a path).
To act intelligently in this setting, a team of robots must
build a shared, semantically meaningful representation of the world,
reliably localize themselves within that representation,
interpret user-specified commands grounded in the environment and robot capabilities,
and produce and execute plans that satisfy the grounded user intent.
In recent years, the research community has addressed these challenges independently, often targeting specific parts of the problem, instead of an integrated system deployed in large-scale environments.
For example, significant progress has been made for constructing 3D scene graphs~\cite{Hughes24ijrr-hydraFoundations,Chang23iros-HydraMulti,Gu24icra-conceptgraphs}, which capture semantics at multiple levels of abstraction to provide a richer understanding of the environment.
The previous work has shown that these representations provide advantages for object-based localization, including hierarchical approaches that utilize scene descriptors~\cite{Hughes24ijrr-hydraFoundations} and open-set approaches with view-invariance~\cite{peterson2025roman}, but have not focused on their use for long-duration, large-scale multi-agent localization and planning.
Similarly, some work has examined the use of  3D scene graphs for planning, but existing efforts have mainly focused on single-robot systems operating in indoor environments~\cite{Ray24isrr-tamp,Rana23corl-sayplan,dai2023optimal}, and their deployment for multi-agent systems in large-scale outdoor environments on real robots has yet to be demonstrated.

\ph{Contributions} In this work, we present two key contributions.
First, we introduce a multi-robot framework for 3D scene graph construction and localization to support task and motion planning (TAMP) given natural language instructions.
Our approach builds a shared scene graph by fusing open-set object-based maps across multiple robots.
We show that this representation supports real-time, view-invariant relocalization (via the object-based map) and planning (via the 3D scene graph), allowing a team of robots to reason about their surroundings and execute complex tasks.
Second, we introduce an approach for hierarchical planning and navigation that translates operator intent expressed in natural language into Planning Domain Definition Language (PDDL)~\cite{Ghallab98-pddl} goals using a Large Language Model (LLM), given context from the scene graph and descriptions of robot capabilities.
We demonstrate the approach with a team of robots in a large-scale, outdoor environment, providing natural language instructions with varying levels of linguistic complexity. Our end-to-end experiment validates the system's ability to reliably ground and execute complex instructions, and we present ablations which further characterize the language grounding procedure and the advantages of our object-based relocalization.
 \begin{figure}[t!]
    \centering
    \includegraphics[width=0.9\columnwidth,trim={0mm 0mm 0mm 0mm},clip]{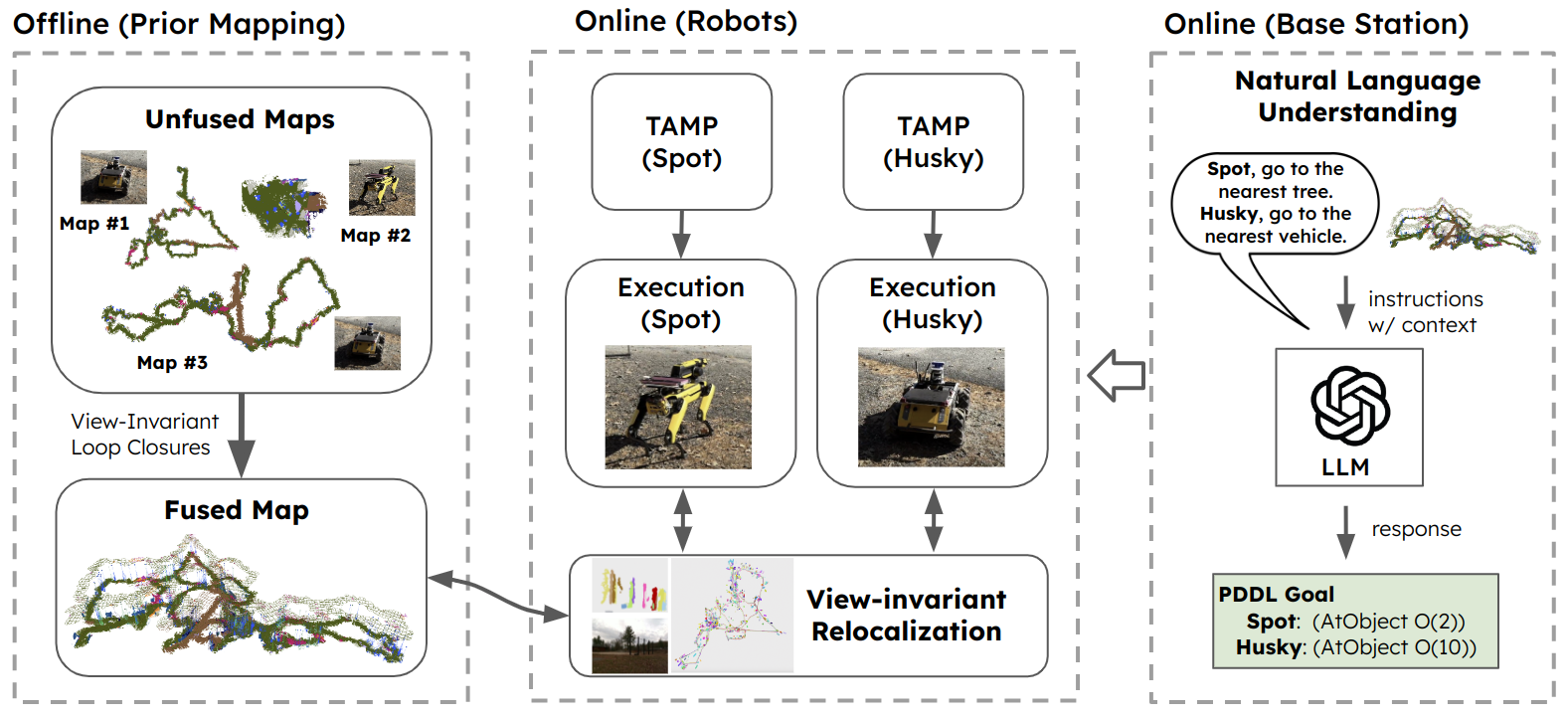}
    \caption{
    {\footnotesize
    The concept of operations consists of a mapping and execution stage. (Left) In the mapping stage, the team of robots construct separate 3D scene graphs and open-set object maps, which are fused into a shared representation. (Right) The user provides natural language instructions, and the user intent is translated into a PDDL goal via an LLM. A PDDL-based planner generates a plan, which the robots executes.
    \label{fig:arch}
    }
    }
    \vspace{-4mm}
\end{figure}
 \vspace{-2mm}
\section{Concept of Operations}
\vspace{-2mm}
The proposed operational procedure consists of an offline mapping stage and an online execution stage.
In the mapping stage, a team of robots is deployed to explore an environment.
In this work, we teleoperate the robots for exploration, but the operational procedure is agnostic to the exploration procedure.
During this process, each robot constructs its own representation of the environment consisting of (i) a 3D scene graph~\cite{Hughes24ijrr-hydraFoundations,Strader24ral-automaticAbstractions} and (ii) an open-set object map~\cite{peterson2025roman}, which are both fused across robots into a consistent shared representation. This shared representation becomes the PDDL planning domain for the execution stage.
In the execution stage, the user provides natural language
instructions describing a task the multi-robot team must complete in the context of the scene graph (\eg{} \emph{``I need a look at the box on the sidewalk and fences along the road.''}).
These instructions are translated into a PDDL goal via an LLM, and a PDDL-based planner generates a plan using this PDDL goal and the scene-graph-defined PDDL domain. The robots then execute this computed plan to complete the task.
This procedure is illustrated in Fig.~\ref{fig:arch}.
 \vspace{-2mm}
\section{Multi-Robot 3D Scene Graph Fusion, Position Estimation, and Relocalization}
\vspace{-2mm}
The goal of the initial mapping phase is to create a shared world representation for a team of robots to estimate their position in while executing the plan, and for humans to reference when providing instructions. In this phase, each robot builds its own scene graph and open-set object map. Then, the scene graphs and object maps are registered and fused by detecting correspondences between maps. The components of the scene graph including objects, places, and regions define the planning domain.
We use Hydra~\cite{Hughes24ijrr-hydraFoundations,Strader24ral-automaticAbstractions} to construct the scene graphs
and ROMAN~\cite{peterson2025roman} to create the maps of open-set objects.
During execution, each robot begins without knowledge of its current pose in the fused map. It creates a new object-based map, which ROMAN aligns to the fused map to localize the robot and enable following plans grounded in the fused scene graph.

\ph{Single-robot Mapping}
3D scene graphs provide a rich, hierarchical structure which is useful for downstream planning and can be built in real-time onboard a robot. Hydra's scene graphs contain a detailed mesh of the environment, an object layer, a places layer representing 3D free space, and a region layer representing larger areas of consistent semantic meaning (\eg{} road or sidewalk). We add a layer of surface places, proposed in~\cite{Ray24isrr-tamp}, to represent navigable 2D space.
The surface places each represent a section of the mesh with consistent semantic labels (\eg{} grass or rocks), which are computed by partitioning the mesh vertices according to their distance from a set of generating place centers. This process repeats incrementally in the Hydra backend, mimicking the behavior of Lloyd's algorithm~\cite{lloyds} for computing a Voronoi partition and resulting in evenly spaced convex polygons. The surface places layer has edges connecting neighboring places, and the resulting graph of places can be used for path planning.
We use ROMAN~\cite{peterson2025roman} to build an open-set object map to detect loop closures and enable relocalization to a prior map.
ROMAN uses FastSAM~\cite{peterson2025roman} to segment color images and then uses depth measurements to track these segments as 3D objects.
Additionally, CLIP embeddings~\cite{radford2021clip} are computed to give objects an open-set semantic descriptor.
ROMAN formulates a robust optimization problem for object matching using geometry, shape, and semantic similarity to associate objects between maps. The key insight is that object-based maps are more view-invariant than image features, enabling loop closures across extreme viewpoint changes.

\ph{Multi-robot Mapping}
We leverage the increased coverage capability of a multi-robot system to create a single large-scale map from the fusion of many single-robot maps.
In the initial mapping phase, each robot is teleoperated through the area and creates online single-robot Hydra scene graphs and ROMAN object maps.
Hydra-Multi~\cite{Chang23iros-HydraMulti} is modified to receive a set of intra- and inter-robot loop closures provided by ROMAN~\cite{peterson2025roman} to jointly optimize each robot pose graph and deformation graph.
Attributes such as the position of each scene graph node are interpolated using the optimized deformation graph.
After interpolation, overlapping nodes of the same class are merged together, resulting in a single fused scene graph that is corrected for odometry drift.
Specific details of the interpolation and fusion can be found in~\cite{Chang23iros-HydraMulti}.
Additionally, the optimized deformation graph is used to correct and combine the object-based maps from ROMAN to enable localizing in the joint map.
Loop closures from ROMAN are computed as follows.
Each single-robot object map is split into a series of submaps containing a collection of objects.
A new submap is created once the robot has traveled {10} {m} away from the center pose of the most recently created submap.
Each object is expressed as a single 3D point in the submap  frame and includes an additional shape descriptor (describing PCA properties of the object point cloud) and its CLIP~\cite{radford2021clip} embedding.
We attempt to perform robust registration of each submap with every other submap to detect loop closures.
Given two submaps, we run the object matching algorithm of ROMAN~\cite{peterson2025roman} to find the set of associations with the greatest consistency.
If two submaps share enough object associations, we compute a loop closure measurement from the transformation that would align the associated objects.
Hydra-Multi receives these putative loop closures, rejects outlier loop closures, and fuses the set of scene graphs into a single multi-robot scene graph.
Finally, the optimized deformation graph computed by Hydra-Multi is used to transform the combined ROMAN object map into the common global coordinate frame.
Additional details on object matching can be found in~\cite{peterson2025roman}.

\ph{Position Estimation and Relocalization}
Task execution during deployment leverages the large-scale fused scene graph to guide language-driven planning.
To navigate in the scene, each robot must relocalize to align their coordinate frame with the fused map frame.
We again use ROMAN~\cite{peterson2025roman} for this, using a similar submap scheme as described above.
Each time a new submap is created, we attempt to associate its objects with those in the submaps from the fused map.
We perform robust transformation averaging to reject outliers and compute the transformation that aligns the robot's local frame to the fused map frame.
Once a robot has relocalized, it can execute plans grounded in the fused map (\ie{} the multi-robot scene graph). \vspace{-2mm}
\section{Language-Grounded Planning}
\vspace{-2mm}
Our system enables a human operator to issue natural language commands to a team of robots based on the information in the fused 3D scene graph.
These commands are translated into PDDL goals using an LLM through in-context learning, and the hierarchical scene graph planner proposed in~\cite{Ray24isrr-tamp} solves the resulting PDDL problem, generating a sequence of actions that can be executed by a robot.
The planner solves a single-agent PDDL problem, and the assignment of separate goals to each agent is handled as part of the language-to-PDDL translation process based on the prompt and scene graph context.

\begin{figure}[t!]
    \centering
    \includegraphics[width=0.99\columnwidth,trim={0mm 0mm 0mm 0mm},clip]{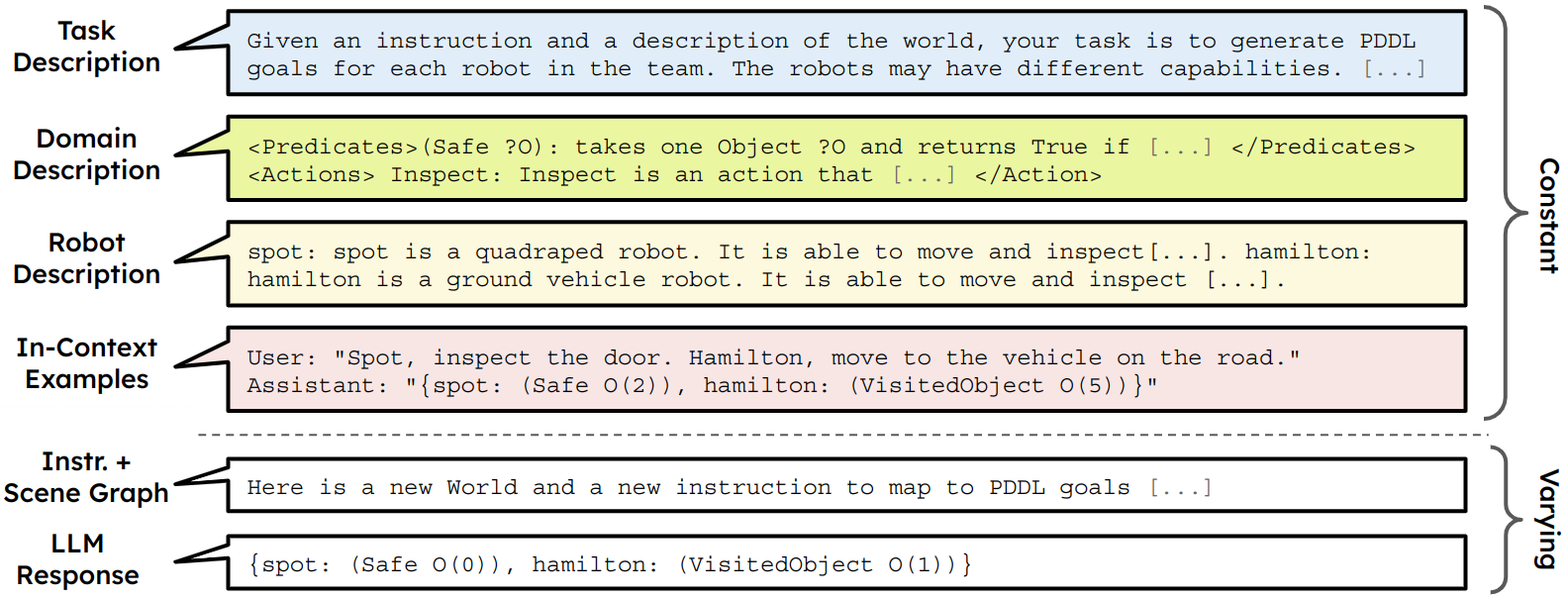}
    \vspace{-2mm}
    \caption{
    {\footnotesize
    Example of LLM prompt and response including snippets for the descriptions of the translation task, PDDL domain, robot capabilities, in-context examples, and instructions for the robots.
    \label{fig:prompt}
    }
    }
    \vspace{-4mm}
\end{figure}

We use PDDL to define our planning problems due to its flexibility in encoding task structure, the straightforward generation of PDDL goals with LLMs, and support for extensions to enable simultaneous task and motion planning~\cite{garrett2020pddlstream}.
Our PDDL domain supports object-centric inspection actions and movement actions that correspond to motion plans through the places layer in the scene graph.
Goal predicates can specify places to visit or avoid and objects to inspect with arbitrary conjunctions, disjunctions, and negations.
The planner uses PDDLStream~\cite{garrett2020pddlstream} to search for both a feasible task skeleton and feasible continuous action parameters.
The scene graph planner outputs a sequence of grounded actions that the robot executes.
To turn natural language commands into PDDL goal propositions grounded in the scene graph, we adapt a neuro-symbolic inference process from prior work~\cite{chen2024autotamp}.
The grounded objects are associated with the referring expressions in the instruction, and the translation is handled by an LLM.
We prompt the LLM with key task information including a text description of the scene graph, a description of robot capabilities, and a set of in-context examples with the expected response format that associates each robot with a PDDL goal. See Fig.~\ref{fig:prompt}.
The text representation of the scene graph contains (i) a list of objects and regions, (ii) their semantic class and position, and (iii) the relationships between objects and regions.
The output, a PDDL goal for each robot, is provided directly to the planning module for planning and execution.

 \vspace{-2mm}
\section{Experiments}
\vspace{-2mm}
In this section, we describe experiments to evaluate the proposed system.
The experiments evaluate (i) multi-robot scene graph fusion and view-invariant relocalization, and (ii) the grounding, planning, and execution of the instructions provided by the operator.
First, we describe the end-to-end experiments where grounding, planning, and execution are completed in sequence using our object-based relocalization.
Second, we complete ablations to better understand the performance of the individual components of the proposed system, which were not practical to complete during the end-to-end experiments.

\begin{figure}[t]
    \centering
    \includegraphics[width=0.99\columnwidth,trim={0mm 0mm 0mm 0mm},clip]{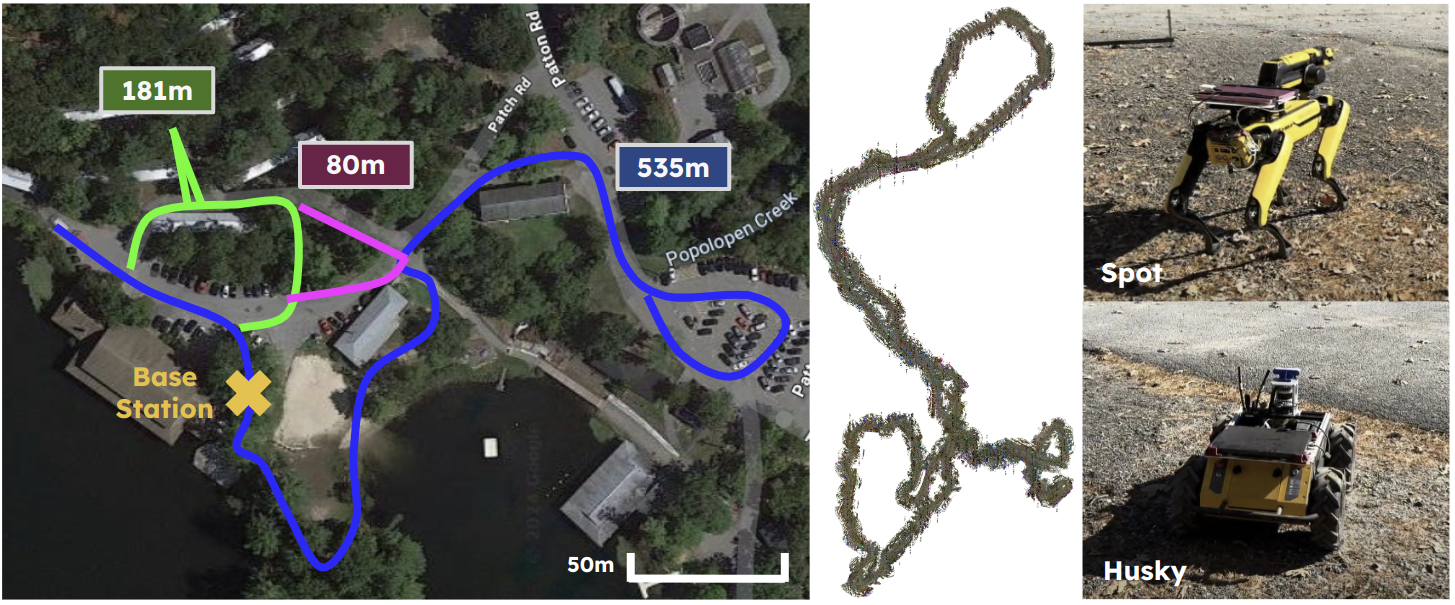}
    \vspace{-2mm}
    \caption{
    {\footnotesize
    (Left) Satellite image of Camp Buckner at West Point, NY, showing the area of operation. The line segments indicate where the robots operate, and the text boxes display their lengths in meters. (Middle) Overhead view of the map constructed from a portion of the blue and green segments. (Right) The platforms used in the experiments.
    }
    \label{fig:setup}}
    \vspace{-8mm}
\end{figure}

\vspace{-2mm}
\subsection{End-to-End Experiments}
\label{sec:endtoend_experiments}

We conduct end-to-end experiments at Camp Buckner in West Point, NY, a large-scale, unstructured outdoor environment
(illustrated in Fig.~\ref{fig:setup}),
using two heterogeneous robots: a wheeled robot (Clearpath Husky) and a quadruped (Boston Dynamics Spot).
Laptops mounted to both the Spot and Husky run the online mapping and low-level autonomy algorithms.
We capture color and depth images using an Intel Realsense D455 camera for the Husky and the built-in forward cameras for the Spot.
We use the internal visual-inertial-kinematic odometry for the Spot and LiDAR odometry~\cite{Reinke22ral-LOCUS2} for the Husky.
The robots use a Silvus radio network to communicate with a centralized natural language interface and planner running at a base station.
The base station displays the fused scene graph to the user, queries an LLM to ground language instructions in the scene graph, and generates a plan for the robots to execute.
These experiments assess the success rate of the system's ability to generate and fuse 3D scene graphs from multiple robots, relocalize within the fused 3D scene graphs, and execute a variety of tasks specified in natural language.

\begin{figure}[t!]
    \centering
    \begin{subfigure}[t]{0.48\columnwidth}
        \centering
        \includegraphics[width=\linewidth, trim={0mm 0mm 0mm 0mm}, clip]{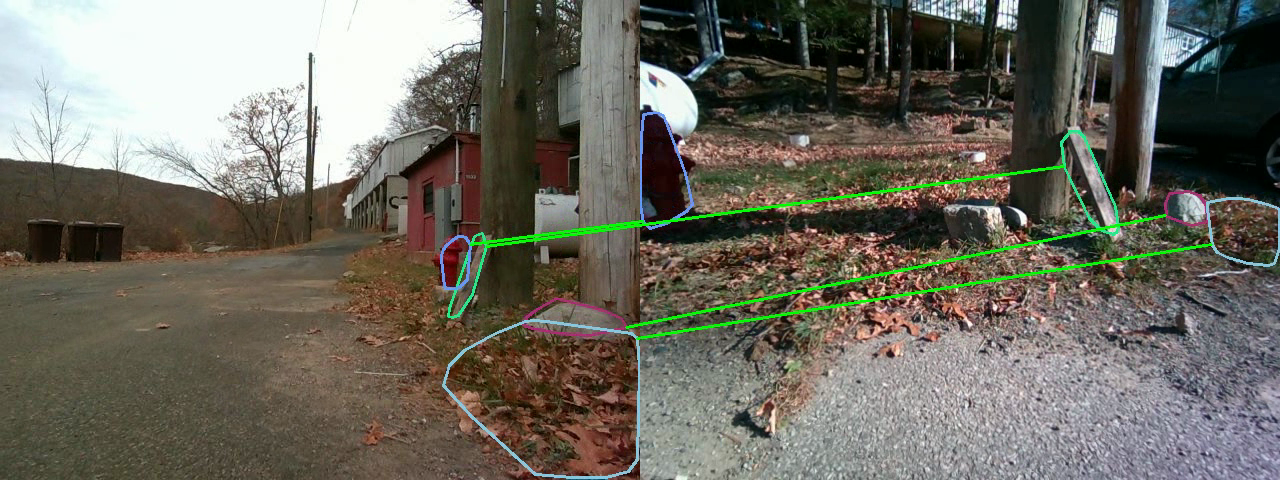}
\end{subfigure}
    \hfill
    \begin{subfigure}[t]{0.48\columnwidth}
        \centering
        \includegraphics[width=\linewidth, trim={0mm 0mm 0mm 0mm}, clip]{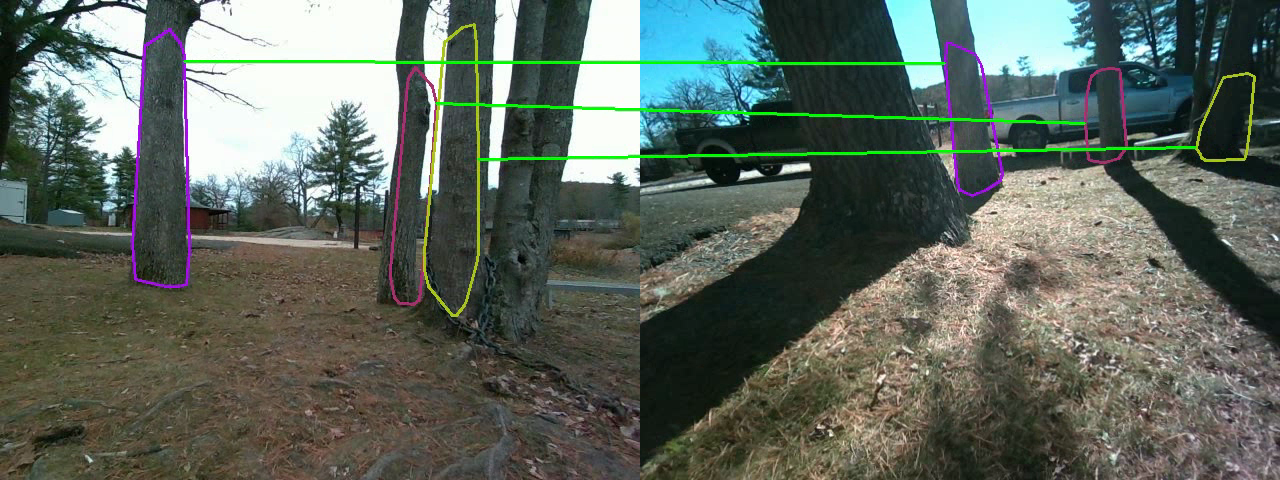}
\end{subfigure}
    \caption{\footnotesize Object matches found by the relocalization system~\cite{peterson2025roman} operating on the Husky front and Spot side cameras.
    Green lines match objects between the two different robot views.
    The images show areas within the operational region at Camp Buckner.}
    \label{fig:object_matching}
\end{figure}

\ph{Position Estimation and Relocalization}
To evaluate the position estimation and relocalization ability provided by the shared scene graph, we considered a portion of the experiment where the Husky and Spot moved along a similar trajectory, roughly 500 meters.
Every 10 meters, we estimated the relative pose between the Husky and Spot cameras (removing any relative pose initialization) using our object-based relocalization.
We used the right-facing camera on Spot and the forward-facing camera on the Husky to test the view-invariance of our relocalization (see Fig.~\ref{fig:object_matching}).
We consider each relocalization instance a success if the estimated relative pose was within 5 meters and 10 degrees of the pseudo ground truth pose computed via pose graph optimization with manually verified loop closures.
The Husky achieved a relocalization success rate of 64\%, while the Spot achieved 65\%, indicating that the object-based relocalization approach maintains consistency across different viewpoints.
Qualitatively, we observe that relocalization performs best where many objects can be observed within the depth range of the cameras, but degrades in areas with no nearby objects (\eg{} stretches of road).

\begin{table*}[t!]
    \centering
    \resizebox{\textwidth}{!}{\begin{tabular}{c ccc ccc ccc}
        \toprule
        & \multicolumn{3}{c}{Easy} & \multicolumn{3}{c}{Medium} & \multicolumn{3}{c}{Hard} \\
        \cmidrule(lr){2-4} \cmidrule(lr){5-7} \cmidrule(lr){8-10}
        & Ground & Plan & Execute & Ground & Plan & Execute & Ground & Plan & Execute \\
        \midrule
        Single-Robot & 100\% & 100\% & 100\% & 100\% & 50\% & 50\% & 75\% & 75\% & 50\% \\
        Multi-Robot  & 100\% & 83\% & 66\% & 100\% & 100\% & 100\% & 100\% & 100\% & 75\% \\
        \bottomrule
    \end{tabular}}
    \vspace{0mm}
    \caption{{\footnotesize Success rates for grounding, planning, and execution across different levels of linguistic complexity (Easy, Medium, Hard) for single-robot (8 trials) and multi-robot (12 trials) experimental setups. The success rate is calculated as the percentage of successful trials.
    A trial is successful if it generates the correct PDDL goal (Ground), produces the correct plan (Plan), and executes the plan to reach the goal (Execute).
    }}
    \label{tab:robot_performance}
    \vspace{-4mm}
\end{table*}

\ph{Mission Execution}
To evaluate planning, we measured success rate across three stages: grounding (\ie{} correctly translating instructions into PDDL goals), planning (\ie{} generating feasible plans from the grounded goals), and execution (\ie{} successfully carrying out the planned actions).
The results are provided in Table~\ref{tab:robot_performance}, which demonstrates the ability to integrate perception and planning in a multi-robot setting across varying difficulties
of commands.
Example commands include: (Easy) \emph{``Spot, go to the closest trash"}, (Medium) \emph{``Spot, inspect the four closest signs"}, and (Hard) \emph{``Spot, check out the two boxes in the shelter and the closest trash, then head over to the pole"}.
While language grounding was mostly successful, we observed a failure for the instruction: \emph{``Spot, either visit the four closest signs or checkout two nearby boxes."}
The object references were grounded to the correct object identifiers for all four signs and the two boxes, but the grounded goal predicates for the boxes had the robot \textit{move} to them, whereas the ground truth PDDL goal was to \textit{inspect} them, as intended by part of the command \emph{``...checkout two nearby boxes"}.

\begin{table*}[b!]
    \centering
    \renewcommand{\arraystretch}{1.0}
    \resizebox{\textwidth}{!}{\begin{tabular}{l @{\hspace{10pt}} l >{\hspace{10pt}} c >{\hspace{10pt}} c >{\hspace{10pt}} c >{\hspace{10pt}} c >{\hspace{10pt}} c} \toprule
        Setting & Method & Object IoU & Precision & Recall & RMSE (m) & Database \\
        \midrule
        \multirow{2}{*}{Hybrid}
        & SALAD~\cite{izquierdo2024salad} & 0.36 ± 0.24 & 0.52 & 0.54 & \textbf{9.40} & 22.4 GB \\
        & \cellcolor{gray!10}ROMAN~\cite{peterson2025roman} & \cellcolor{gray!10}\textbf{0.40 ± 0.25} & \cellcolor{gray!10}\textbf{0.54} & \cellcolor{gray!10}\textbf{0.55} & \cellcolor{gray!10}10.36 & \cellcolor{gray!10}\textbf{17.2 MB} \\
        \midrule
        \multirow{2}{*}{Outdoor}
        & SALAD~\cite{izquierdo2024salad} & 0.34 ± 0.20 & 0.55 & 0.57 & \textbf{5.67} & 15.0 GB \\
        & \cellcolor{gray!10}ROMAN~\cite{peterson2025roman} & \cellcolor{gray!10}\textbf{0.36 ± 0.19} & \cellcolor{gray!10}\textbf{0.63} & \cellcolor{gray!10}\textbf{0.68} & \cellcolor{gray!10}6.61 & \cellcolor{gray!10}\textbf{9.8 MB} \\
        \bottomrule
    \end{tabular}}
    \vspace{0mm}
    \caption{{\footnotesize
    Comparison of ROMAN and SALAD with respect to the RMSE of ATE, object pose accuracy (IoU, Precision, Recall), and the memory usage for their databases.
}}
    \label{tab:fusion_results}
    \vspace{-4mm}
\end{table*}

\vspace{-2mm}
\subsection{Ablations}
\ph{Scene Graph Fusion}
To understand the performance of the scene graph fusion beyond the qualitative results from the end-to-end experiments, we examined the accuracy of our multi-robot scene graph fusion on the Kimera-Multi dataset~\cite{Tian23iros-kimeraMulti}.~\footnote{We do not complete the ablations during the real-time experiments at Camp Buckner due to the lack of ground truth.}
Specifically, we used the \emph{hybrid} and \emph{outdoor} sequences, and for each sequence, we used the data for the \emph{acl\_jackal2}, \emph{sparkal1}, and \emph{sparkal2} robots.
To showcase the benefits of ROMAN, we compared against a competitive visual loop-closure baseline that uses SALAD~\cite{izquierdo2024salad} for place descriptors, SuperPoint~\cite{detone2018superpoint} for local features, and LightGlue~\cite{lindenberger2023lightglue} for feature matching.
We report the Root Mean Square Error (RMSE) of the Absolute Trajectory Error (ATE) of the optimized robot trajectories, along with the precision, recall, and Intersection over Union (IoU) of the objects in the fused scene graph as defined in~\cite{Chang23iros-HydraMulti} with a tolerance of $5 m$.
The results are shown in Table~\ref{tab:fusion_results}.
These metrics validate the improved accuracy of ROMAN when integrated with Hydra-Multi as compared to traditional visual place recognition based on visual place descriptors and feature matching.
While loop closures from ROMAN resulted in a slightly higher ATE RMSE, the optimized scene graphs tended to have higher global consistency, resulting in improvements in object IoU, precision, and recall.
Additionally, ROMAN requires minimal data storage and transmission for computing loop closures.
We observed better performance when ROMAN has access to gravity-aligned odometry (\eg from IMU), as demonstrated by the ATE RMSE result of $6.91 m$ reported in~\cite{peterson2025roman} on the same $hybrid$ data sequences.
Furthermore, by experimental design, much of the trajectory overlap occured when robots were facing similar directions where visual feature matching methods excel.

\begin{table*}[t!]
    \centering
    \renewcommand{\arraystretch}{1.1}  \rowcolors{2}{gray!10}{white}      \resizebox{\textwidth}{!}{\begin{tabular}{l l}
        \toprule
        \textbf{Linguistic Category} & \textbf{Example Instruction} \\
        \midrule
        Direct scene graph concepts     & Spot, inspect objects 39, 55, 395, and 397. \\
        Unambiguous action command      & Spot, inspect the trash. \\
        Unambiguous goal command        & Spot, be by the window. \\
Co-reference resolution         & The boxes are unsafe. Spot, inspect one of them. \\
        Spatial relation disambiguation & Spot, inspect the box closest to the bag. \\
        Region-level disambiguation     & Spot, checkout both signs in the parking lot. \\
        Ambiguous role assignment       & One of you go to the window. The other should inspect the trash. \\
        \bottomrule
    \end{tabular}}
    \vspace{0.0em}
    \caption{{\footnotesize Examples of linguistic categories with representative instructions. Experiments were completed using 10 instructions per category.}}
    \label{tab:linguistic-categories}
    \vspace{-4mm}
\end{table*} \begin{table}[h!]
    \centering
    \renewcommand{\arraystretch}{1.1}
    \vspace{2mm}
    \resizebox{\columnwidth}{!}{\rowcolors{2}{gray!10}{white}
    \begin{tabular}{
        l
        >{\hspace{6pt}}c<{\hspace{6pt}}
        >{\hspace{6pt}}c<{\hspace{6pt}}
        >{\hspace{6pt}}c<{\hspace{6pt}}
        >{\hspace{6pt}}c<{\hspace{6pt}}
        >{\hspace{6pt}}c<{\hspace{6pt}}
        >{\hspace{6pt}}c<{\hspace{6pt}}
        >{\hspace{6pt}}c<{\hspace{6pt}}
        >{\hspace{6pt}}c<{\hspace{6pt}}
    }
        \toprule
        Model & DC & UA-Cmd & UG-Cmd & CR & SR-D & RL-D & AR-A & Total \\
        \midrule
        GPT-4o               & 9  & 7  & 8  & \colorbox{yellow!20}{10} & 2 & 4 & 9  & 49 / 70 \\
        GPT-4o-mini          & 7  & 4  & 5  & 4  & 2 & 2 & 4  & 28 / 70 \\
        GPT-4.1              & 8  & \colorbox{yellow!20}{9}  & \colorbox{yellow!20}{9}  & 9  & 4 & \colorbox{yellow!20}{5} & \colorbox{yellow!20}{10} & \colorbox{yellow!20}{54 / 70} \\
        GPT-4.1-mini         & 9  & 6  & \colorbox{yellow!20}{9}  & 8  & 2 & 4 & 5  & 43 / 70 \\
        GPT-4.1-nano         & 5  & 1  & 0  & 2  & 0 & 0 & 0  & 8 / 70 \\
        Claude 3.7 Sonnet v1 & \colorbox{yellow!20}{10} & 6  & 8  & 8  & 4 & 4 & 9  & 49 / 70 \\
        Claude 3.5 Sonnet v2 & 9  & 7  & 7  & 7  & \colorbox{yellow!20}{5} & 1 & 6  & 42 / 70 \\
        Claude 3.5 Haiku v1  & 9  & 7  & 6  & 5  & 4 & \colorbox{yellow!20}{5} & 4  & 40 / 70 \\
        \bottomrule
    \end{tabular}
    }
    \vspace{1mm}
    \caption{\footnotesize Number of correctly grounded commands per category (10 commands per category). Abbreviations: DC = Direct Concepts, UA-Cmd = Unambiguous Action Command, UG-Cmd = Unambiguous Goal Command, CR = Co-reference Resolution, SR-D = Spatial-relation Disambiguation, RL-D = Region-level Disambiguation, AR-A = Ambiguous Role Assignment. The highest score in each column is highlighted.}
    \label{tab:nlp_results}
    \vspace{-6mm}
\end{table}

\ph{Language Grounding}
To better characterize the performance of the language grounding procedure, we provide additional offline experiments over a larger set of instructions applied to the fused 3D scene graph from Camp Buckner.
We constructed a dataset consisting of instructions and their corresponding ground truth PDDL goals assigned to different robots per instruction.
We evaluated our inference process for grounding natural language commands to PDDL goals for seven distinct linguistic categories (see Table~\ref{tab:linguistic-categories}), which emphasized different dimensions of the language-to-goal generation problem.
The success rate for grounding commands in these categories to PDDL goals for a range of model sizes is shown in Table~\ref{tab:nlp_results}. Notably, all models struggled on both the spatial relation and region-level disambiguation categories. We attribute this to both LLMs not being well suited for numerical evaluations (spatial relations) and the size of the scene graph in the prompt (object-region associations). Investigating better techniques for LLMs to reason about scene graphs is an aim for future work.

 \vspace{-2mm}
\section{Related Work}
\label{sec:related_works}
\vspace{-2mm}

Introduced in~\cite{armeni20193d} and extended in~\cite{rosinol2021ijrr}, 3D scene graphs provide a unified structure for storing and reasoning about spatial and semantic information in a scene, and can be constructed in real time~\cite{Hughes24ijrr-hydraFoundations,bavle2023s}.
Recent works have demonstrated 3D scene graphs can be leveraged to infer information about higher-level abstractions, such as room labels or region groupings~\cite{Strader24ral-automaticAbstractions}.
Additional work from \textcite{Chang23iros-HydraMulti} has expanded these approaches to multi-agent systems, which we extend in this work. 3D scene graphs have also been extended to the open-vocabulary setting using language-aligned image embeddings such as CLIP~\cite{radford2021clip}, and several works have shown this is useful for grounding object search tasks~\cite{Gu24icra-conceptgraphs, maggio2024clio, werby2024hierarchical}.

To support planning in 3D scene graphs, several methods derive structured planning domains from these representations. \textcite{Agia22corl-Taskography} convert scene graphs to PDDL domains to exploit structure to prune the planning problem.
\textcite{dai2023optimal} formulate LTL problems from scene graph symbols, and similarly, \textcite{Ray24isrr-tamp} demonstrate efficient TAMP using scene graphs to identify when symbols can be removed without compromising feasibility. We build upon these works, translating natural language to PDDL for TAMP, using the planner from~\cite{Ray24isrr-tamp}.
While LLMs can act as planner in certain domains~\cite{huang2022languagemodelszeroshotplanners,ahn2022icanisay,Rana23corl-sayplan}, they often struggle with realistic TAMP problems~\cite{silver2022pddl, valmeekam2022large}.
In~\cite{chen2024autotamp}, natural language is translated into STL as an intermediate representation and solved with a traditional planner, improving efficiency for complex problems.
Other approaches translate language into LTL~\cite{dai2023optimal}, PDDL goals ~\cite{xie2023translatingnaturallanguageplanning} or full PDDL problems ~\cite{liu2023llmpempoweringlargelanguage}.

To constrain drift and build a shared representation, a robot must detect loop closures with observations across robots.
This is usually completed in two stages: place recognition and relative pose estimation.
Typically, this involves extracting local and global features~\cite{rublee2011orb, galvez2012bags}, which have recently showed improved results enabled by deep learning~\cite{sarlin2020superglue,izquierdo2024salad}.
However, these methods suffer from sensitivity to sensor noise, appearance change, and viewpoint shifts.
In contrast, object-level methods tend to be more robust against appearance changes and can be more compact and lightweight~\cite{gawel2018x, liu2024slideslam, wang2024goreloc}.
For example, ROMAN \cite{peterson2025roman}, which we use in this work, builds on CLIPPER \cite{lusk2024clipper} and aims to find large sets of consistent object-level associations.

 \vspace{-2mm}
\section{Conclusions}
\vspace{-2mm}
In this work, we presented a multi-robot system that integrates 3D scene graph generation with language-grounded task and motion planning to execute natural language instructions.
We demonstrate the effectiveness of our approach through real-time experiments in a large-scale environment, as well as ablations on datasets targeting different components of the system.

While the results are promising, this work opens several directions for future research.
For example, the robots are currently constrained to operating within the previously fused map, which limits the system from handling exploration tasks or adapting to changes in the environment.
Further, the language grounding does not yet exploit the open-set nature of the object-based map to support more open-ended tasks.
These are exciting directions for future research, with the potential to enable robots to complete a wider variety of tasks.

\def\bibfont{\small}
\printbibliography

\end{document}